\documentclass[runningheads]{llncs}

\usepackage{graphicx}
\usepackage{amsmath}
\usepackage{amssymb}
\usepackage{amsfonts}
\usepackage{booktabs}
\usepackage{algorithm}
\usepackage{algorithmic}
\usepackage{soul}
\usepackage{url}
\usepackage[hidelinks]{hyperref}
\usepackage[utf8]{inputenc}
\usepackage[switch]{lineno}
\usepackage[protrusion=true,expansion=false]{microtype}
\hypersetup{
  pdftitle={Teacher-Guided Causal Interventions for Image Denoising: Orthogonal Content--Noise Disentanglement in Vision Transformers},
  pdfauthor={Kuai Jiang, Zhaoyan Ding, Guijuan Zhang, Dianjie Lu, Zhuoran Zheng},
  pdfkeywords={Image Denoising, Causal Intervention, Vision Transformer, Disentanglement}
}

\begin{document}

\title{Teacher-Guided Causal Interventions for Image Denoising: Orthogonal Content--Noise Disentanglement in Vision Transformers}
\titlerunning{Teacher-Guided Causal Interventions for Image Denoising}

\author{%
\small
\mbox{%
Kuai Jiang\inst{1}\thanks{Kuai Jiang and Zhaoyan Ding contributed equally to this work.}
\and Zhaoyan Ding\inst{1}\ensuremath{^{\star}}
\and Guijuan Zhang\inst{2}
\and Dianjie Lu\inst{2}
\and Zhuoran Zheng\inst{3}%
}}
\authorrunning{K. Jiang et al.}
\institute{China University of Mining and Technology\\
\email{\{kuaijiang,dingzhaoyan\}@cumt.edu.cn}
\and
Shandong Normal University\\
\email{\{zhangguijuan,ludianjie\}@sdnu.edu.cn}
\and
Qilu University of Technology\\
\email{zhengzr@njust.edu.cn}}

\maketitle
\vspace{-0.10cm}

\begin{abstract}
Conventional image denoising models often inadvertently learn spurious correlations between environmental factors and noise patterns. Moreover, due to high-frequency ambiguity, they struggle to reliably distinguish subtle textures from stochastic noise, resulting in over-removed details or residual noise artifacts. We therefore revisit denoising via causal intervention, arguing that purely correlational fitting entangles intrinsic content with extrinsic noise, which directly degrades robustness under distribution shifts. Motivated by this, we propose the Teacher-Guided Causal Disentanglement Network (TCD-Net), which explicitly decomposes the generative mechanism via structured interventions on feature spaces within a Vision Transformer framework. Specifically, our method integrates three key components: (1) An Environmental Bias Adjustment (EBA) module projects features into a stable, de-centered subspace to suppress global environmental bias (de-confounding). (2) A dual-branch disentanglement head employs an orthogonality constraint to force a strict separation between content and noise representations, preventing information leakage. (3) To resolve structural ambiguity, we leverage Nano Banana Pro, Google's reasoning-guided AI image generation model, to guide a causal prior, effectively pulling content representations back onto the natural-image manifold. Extensive experiments demonstrate that TCD-Net outperforms mainstream methods across multiple benchmarks in both fidelity and efficiency, achieving a real-time speed of 104.2 FPS on a single RTX 5090 GPU.

\end{abstract}

\newpage

\section{Introduction} 
\begin{figure}[t]
  \centering
  \includegraphics[width=0.95\textwidth]{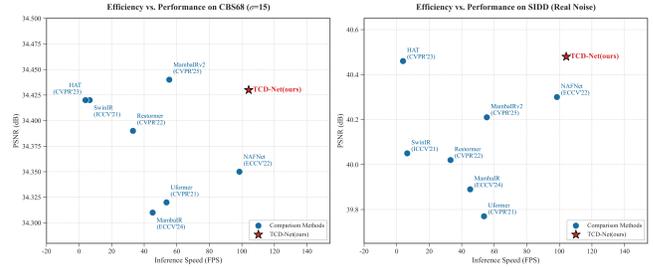}
  \caption{\textbf{Efficiency--performance trade-off.} PSNR vs.\ FPS on CBSD68 ($\sigma{=}15$) and SIDD. TCD-Net achieves the best speed--quality trade-off among competing methods.}
  \label{fig:tradeoff}
\end{figure}

Image denoising is a fundamental step in computational photography and low-level vision, serving as a prerequisite for downstream perception and prediction. Yet, denoising remains intrinsically ill-posed: the observation is jointly shaped by intrinsic scene content and extrinsic corruption sources (e.g., sensor noise, ISP pipelines, illumination), while both fine textures and noise often appear as high-frequency signals. Consequently, data-driven denoisers may exploit spurious correlations between environment factors and noise patterns, leading to over-smoothing or residual artifacts (see Fig.~\ref{fig:tradeoff}).

Classical denoisers mitigate ill-posedness with hand-crafted priors such as BM3D \cite{dabov2007bm3d}, while learning-based methods learn data-driven mappings (e.g., DnCNN \cite{zhang2017dncnn} and lightweight NAFNet \cite{chen2022nafnet}). For real photographs, complex camera/ISP noise motivates evaluation on SIDD/DND \cite{abdelhamed2018sidd,plotz2017dnd} and noise-aware modeling. Self/zero-shot schemes such as Deep Image Prior further reduce reliance on clean targets \cite{ulyanov2018dip}. However, without explicit constraints on the generation process, many denoisers still entangle textures with stochastic noise via correlation-based fitting.

Transformers are increasingly adopted for image restoration. ViT demonstrates that global self-attention over patch tokens can be effective \cite{dosovitskiy2021vit}. Hierarchical designs enable scalable dense prediction \cite{liu2021swin}, and sophisticated restoration backbones, exemplified by Restormer and HAT, have demonstrated exceptional efficacy across denoising and related tasks \cite{zamir2022restormer,chen2023hat}. Yet, without explicit structural constraints, even strong backbones may learn shortcuts tied to nuisance cues and degrade under distribution changes.

Beyond attention, state-space models provide an efficient alternative. Mamba introduces selective state spaces with linear-time sequence modeling \cite{gu2023mamba}, and recent restorers adapt this paradigm for image restoration, including MambaIR \cite{guo2024mambair}. Despite favorable efficiency and global receptive fields, SSM-based restorers can be sensitive to 2D inductive biases (e.g., scan/flatten order) and may aggregate local details less effectively than carefully designed attention or convolution modules, especially under real-noise and domain-shift settings.

In this work, we revisit denoising from a causal intervention perspective. Causal inference distinguishes correlation from causation by modeling data generation via structural causal models (SCMs) and reasoning about interventions \cite{pearl2009causality}. For denoising, the noisy observation arises from intrinsic content factors and extrinsic noise/environment factors; purely correlational fitting can therefore mix texture cues with nuisance variations in a single representation, harming robustness. Causal representation learning further argues that disentangling underlying causal factors is key to generalization and transfer \cite{scholkopf2021towardcausalrep}. This underscores the necessity of developing denoisers as structured interventions that suppress environment-induced bias, enforce factor separation, and improve identifiability.

Based on this causal formulation, we propose \textbf{TCD-Net} (Teacher-Guided Causal Disentanglement Network), a Vision Transformer denoiser that performs structured interventions to explicitly disentangle content and noise. We design an \textbf{Environmental Bias Adjustment (EBA)} module, a projection-and-restoration operator that removes token-wise bias and reinjects corrected features, to deconfound global environment shifts. We further design a \textbf{dual-head disentanglement} architecture to jointly predict the restored image and an explicit noise map, and enforce an \textbf{orthogonality constraint} between content/noise subspaces with \textbf{strong noise supervision} to prevent cross-branch leakage and anchor the noise branch. Finally, we incorporate a \textbf{teacher-guided causal prior} based on \textbf{Google Nano Banana Pro (NBP)}\cite{googledeepmind2025nanobananapro}, which provides strong \emph{zero-shot} restoration guidance and more natural high-frequency details \cite{zuo2025nanobanana}. To avoid over-trusting potentially hallucinated textures, we distill NBP only during training via a \emph{feature-level} perceptual regularizer (instead of strict pixel matching), complementing perceptual supervision \cite{hinton2015distill,johnson2016perceptual}. Moreover, since ViT-style absolute positional embeddings may break translation equivariance and rely on interpolation under resolution changes, we adopt resolution-adaptive conditional positional encoding (CPE) \cite{chu2023cpvt} to mitigate positional representation shifts and improve robustness under distribution shifts.
\vspace{-0.20cm}
\paragraph{Contributions.}
(1) We introduce a causal-intervention formulation for image denoising and instantiate it as TCD-Net, explicitly disentangling content and noise within a Vision Transformer.
(2) We propose Environmental Bias Adjustment (EBA)-based deconfounding and orthogonal content--noise subspace constraints with strong noise supervision for anchoring.
(3) We integrate a Google Nano Banana Pro (NBP)-guided causal prior to improve identifiability and perceptual fidelity.
(4) Extensive experiments and systematic ablations validate the effectiveness and efficiency of each component.

%%%%%%%%%%%%%%%%%%%%%%%%%%%%%%%%%%%%%%%%%%%%%%%%%%%%%%%%%%%%%%%%%%%%%%%%%%%%%%
% Related Work
%%%%%%%%%%%%%%%%%%%%%%%%%%%%%%%%%%%%%%%%%%%%%%%%%%%%%%%%%%%%%%%%%%%%%%%%%%%%%%
\section{Related Work}
\subsection{Denoising Backbones: CNN/Transformer/SSM}
Image denoising has progressed from hand-crafted priors such as BM3D \cite{dabov2007bm3d} to learning-based mappings exemplified by residual CNNs (DnCNN) \cite{zhang2017dncnn} and efficient modern designs (NAFNet) \cite{chen2022nafnet}, while real camera noise benchmarks (SIDD/DND) expose the synthetic-to-real gap \cite{abdelhamed2018sidd,plotz2017dnd}. Recent restorers increasingly rely on global context modeling, including Transformer-based architectures (Restormer, HAT) \cite{zamir2022restormer,chen2023hat} and resolution-stable conditional positional encoding (CPE) \cite{chu2023cpvt}. Beyond attention, state-space models provide linear-time global mixing via Mamba \cite{gu2023mamba} and restoration adaptations such as MambaIR \cite{guo2024mambair}. Despite these advances, stronger backbones or conditioning alone do not resolve the core ambiguity: without explicit structural constraints, models may still entangle textures with stochastic noise and overfit nuisance cues, degrading robustness under distribution shifts.

\subsection{Generative/Teacher Priors and Causal Deconfounding}
Generative priors (e.g., diffusion models) enhance perceptual realism but often require iterative sampling and are sensitive to mismatched noise processes, raising deployment and stability concerns. From a causal viewpoint, denoising can be framed with SCMs and interventions \cite{pearl2009causality}, where a key direction is to remove environment-induced confounders and separate content from nuisance noise for improved OOD robustness. In parallel, teacher guidance injects semantic priors via distillation \cite{hinton2015distill}; we instantiate this idea with a \emph{Google Nano Banana Pro} (NBP)-guided prior \cite{zuo2025nanobanana}, which can be distilled once during training and keeps test-time inference single-pass, better fitting real-time denoising. To avoid over-trusting potentially hallucinated details, we distill NBP at the feature level and combine it with explicit factorization (a noise-map branch) and orthogonal subspace constraints, yielding a causal-factorized and efficient denoiser rather than a sampling-based generator.

%%%%%%%%%%%%%%%%%%%%%%%%%%%%%%%%%%%%%%%%%%%%%%%%%%%%%%%%%%%%%%%%%%%%%%%%%%%%%%
% Method
%%%%%%%%%%%%%%%%%%%%%%%%%%%%%%%%%%%%%%%%%%%%%%%%%%%%%%%%%%%%%%%%%%%%%%%%%%%%%%
\section{Method}\label{sec:method}
\begin{figure}[t]
  \centering
  \includegraphics[width=0.95\textwidth]{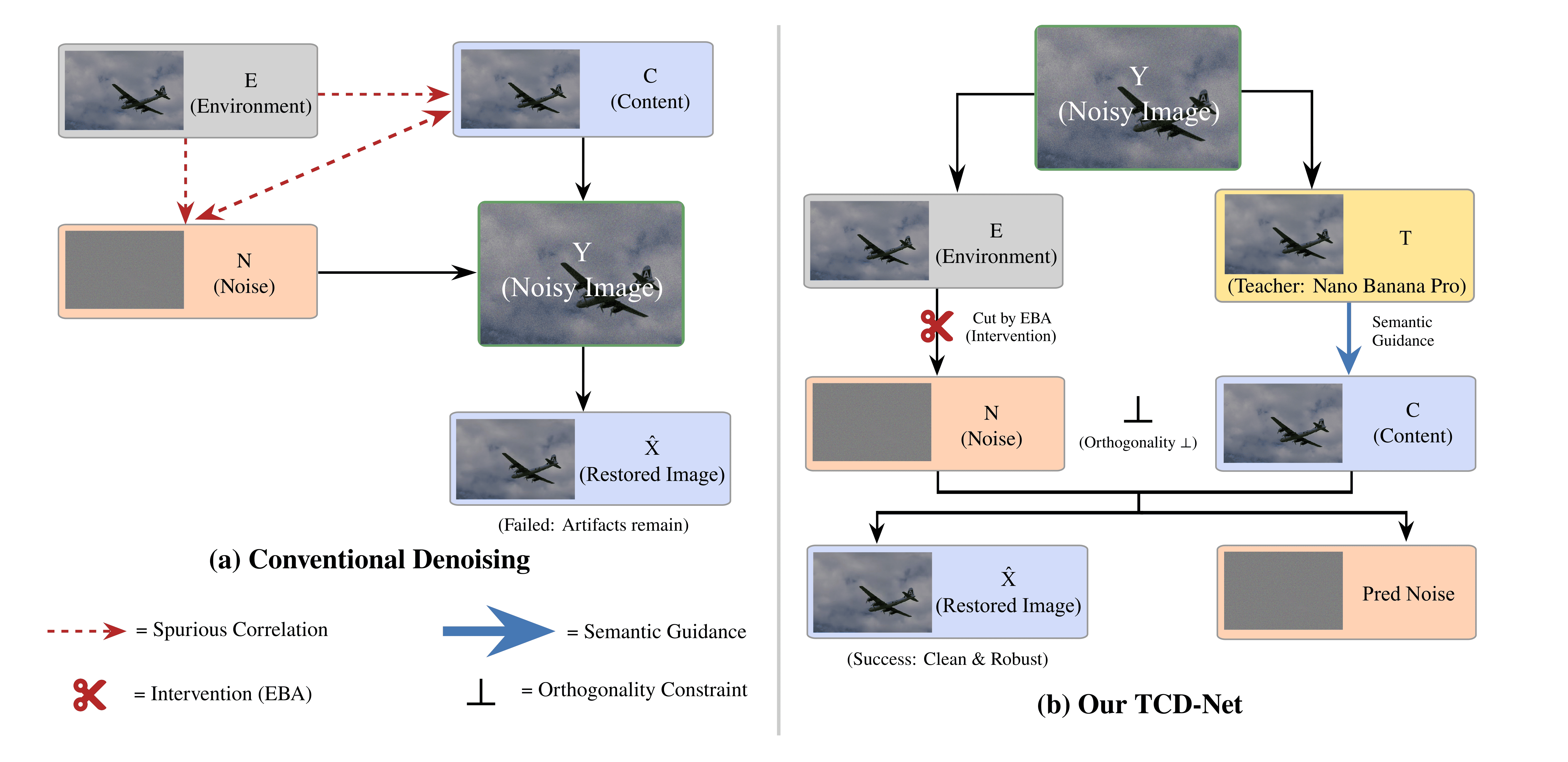}
  \caption{\textbf{Conceptual comparison between conventional denoising and TCD-Net.}
  \textbf{(a) Conventional denoising} is prone to spurious content--noise correlations induced by environmental factors $E$, leaving residual artifacts.
  \textbf{(b) TCD-Net} performs causal intervention via \emph{EBA}, incorporates \emph{teacher semantic guidance}, and enforces an \emph{orthogonality constraint} to decouple content and noise for cleaner restoration.}
  \label{fig:concept}
\end{figure}

We instantiate the causal-intervention view in Introduction as a concrete Vision Transformer denoiser, \textbf{TCD-Net}, whose core goal is to \emph{explicitly separate intrinsic content from extrinsic noise} and to \emph{stabilize} this separation under environment/resolution shifts.

\subsection{Causal View and Overall Architecture}
Let $Y \in \mathbb{R}^{H\times W\times 3}$ be the observed noisy image. We model $Y$ as jointly generated by intrinsic \textbf{content} $C$ (scene structures/textures) and extrinsic \textbf{noise} $N$ (sensor/ISP corruption), under an \textbf{environment} factor $E$ (illumination, gain, pipeline) that induces domain shift. A generic structural causal model (SCM) can be concisely written as:
\begin{equation}
\label{eq:scm}
C := f_C(U_C),\quad E := f_E(U_E),\quad N := f_N(C,E,U_N),\quad Y := f_Y(C,N,E),
\end{equation}
where $U_\cdot$ are exogenous variables. In denoising, we observe $Y$ but must infer $C$. Given \eqref{eq:scm}, learning content $C$ from the observation $Y$ by correlational fitting can spuriously couple $C$ and noise $N$, especially in high-frequency regions where textures and noise overlap. We therefore design TCD-Net as structured interventions: EBA removes $E$-induced bias, a dual-branch head factorizes content/noise, an orthogonality constraint cuts leakage, explicit noise supervision anchors $N$, and an NBP-guided prior regularizes the content manifold.

Given an input $Y$, TCD-Net predicts both a restored image $\hat{X}$ and an explicit noise map $\hat{N}$:
\begin{equation}
\label{eq:dualout}
(\hat{X},\hat{N}) = \mathcal{F}_\theta(Y).
\end{equation}
The architecture is a ViT-style denoiser that directly predicts clean content (on-manifold) while simultaneously modeling noise (off-manifold). Concretely, a Transformer encoder extracts mixed features $Z_{\text{all}}$, then a dual-branch head disentangles them into content features $Z_c$ and noise features $Z_n$, which are decoded into $\hat{X}$ and $\hat{N}$, respectively. 
The five causal interventions are implemented as: EBA-based de-confounding, dual-branch factorization, orthogonal subspace constraints, strong noise anchoring, and NBP-guided causal priors.

\begin{figure}[t]
  \centering
  \includegraphics[width=0.95\textwidth]{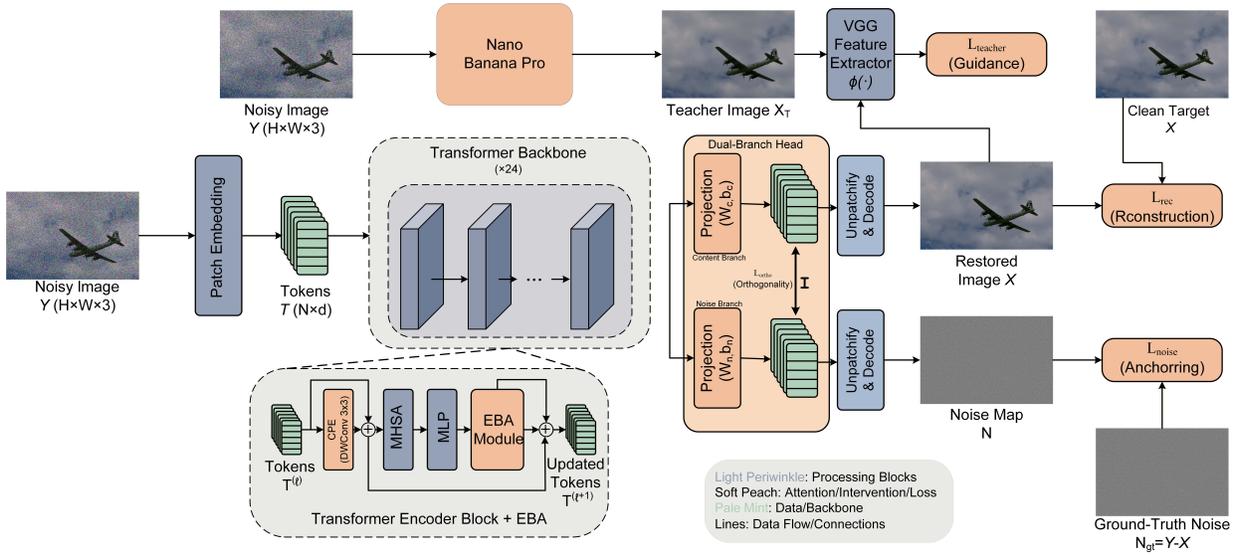}
  \caption{\textbf{Overview of TCD-Net.} A ViT backbone with EBA feeds a dual-branch head to predict the restored image $\hat{X}$ and noise map $\hat{N}$, trained with orthogonality, noise anchoring, and teacher guidance.}
  \label{fig:arch}
\end{figure}

Fig.~\ref{fig:concept} illustrates why purely correlational denoising entangles content with noise, and how our causal intervention design explicitly breaks such spurious paths.
The overall pipeline of TCD-Net is summarized in Fig.~\ref{fig:arch}.

\subsection{Transformer Backbone with Resolution-Stable Positional Encoding}
We follow a ViT patch-tokenization pipeline. Given $Y\in\mathbb{R}^{H\times W\times 3}$, a convolutional patch embedding with patch size $p$ produces a token grid:
\begin{equation}
\label{eq:patch_embed}
T = \text{PatchEmbed}(Y) \in \mathbb{R}^{N \times d},\quad N=\frac{H}{p}\cdot\frac{W}{p}.
\end{equation}
We use a class token only for compatibility with standard ViT implementations and discard it for dense prediction. To mitigate positional representation shift under resolution changes, we use a hybrid positional scheme: (i) interpolated absolute embeddings, where the embedding grid is bicubically interpolated to match $(H/p,W/p)$ when needed; and (ii) conditional positional encoding (CPE), where each block injects a depth-wise $3\times 3$ convolutional positional signal computed from the \emph{current} token features. Let $T^{(\ell)}$ be tokens at block $\ell$ reshaped to a feature map $\mathcal{R}(T^{(\ell)})\in\mathbb{R}^{d\times H/p \times W/p}$. Then
\begin{equation}
\label{eq:cpe}
T^{(\ell)} \leftarrow T^{(\ell)} + \text{Flatten}\!\Big(\text{DWConv}_{3\times 3}\big(\mathcal{R}(T^{(\ell)})\big)\Big).
\end{equation}
Each block then applies MHSA and an MLP, followed by our EBA de-confounding module to further improve invariance to environment shifts.

\subsection{Causal Interventions: De-confounding, Disentanglement, and NBP Guidance}

\subsubsection{Environmental Bias Adjustment (EBA) de-confounding.}
\begin{figure}[t]
  \centering
  \includegraphics[width=0.75\columnwidth]{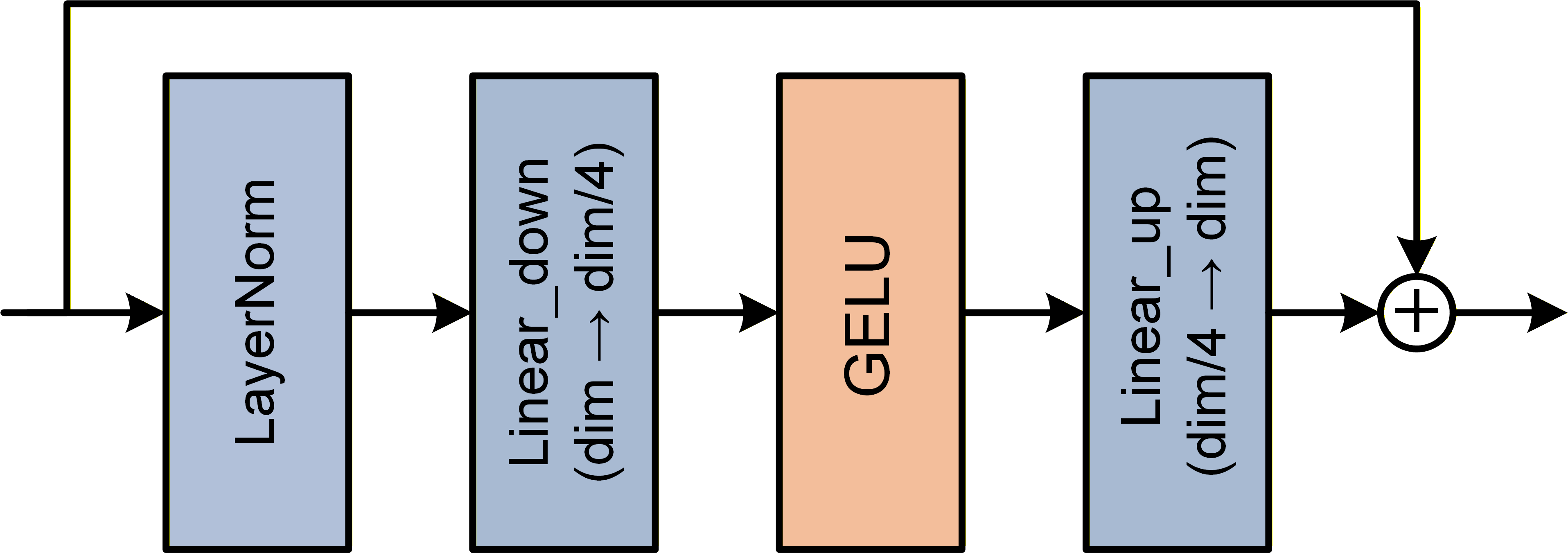}
  \caption{\textbf{EBA module.} LayerNorm + bottleneck MLP with residual projection to suppress environment-induced bias and stabilize token representations.}
  \label{fig:eba}
\end{figure}

Real noisy images often contain global appearance shifts (e.g., illumination/color temperature) that act as
environment-induced confounders $E$. We implement de-confounding with our \textbf{Environmental Bias
Adjustment (EBA)} module, embedded at the end of each Transformer block. For a token feature
$t\in\mathbb{R}^{d}$, EBA performs explicit de-centering, projection via a
bottleneck MLP, and affine restoration via a residual connection:
\begin{equation}
\label{eq:eba}
\mu(t) = \frac{1}{d}\mathbf{1}^\top t, \quad \tilde{t} = t - \mu(t)\mathbf{1}, \quad \text{EBA}(t) = t + W_2\,\sigma\!\big(W_1\,\text{LN}(\tilde{t})\big),
\end{equation}
where $\text{LN}(\cdot)$ is LayerNorm, $\sigma(\cdot)$ is GELU, and $W_1,W_2$ are learnable projections. Removing the per-token centroid suppresses global bias, while the projection stabilizes features toward a more robust subspace.
The concrete operator structure of EBA is shown in Fig.~\ref{fig:eba}.

\subsubsection{Dual-branch content--noise disentanglement and orthogonality.}
Let $Z_{\text{all}} \in \mathbb{R}^{N\times d}$ denote the final encoder patch-token features. We explicitly parameterize two branches:
\begin{equation}
\label{eq:split}
Z_c = Z_{\text{all}} W_c + b_c,\qquad
Z_n = Z_{\text{all}} W_n + b_n,
\end{equation}
where $(W_c,b_c)$ and $(W_n,b_n)$ are learned linear projections. Each branch predicts patch-wise RGB values, then we \emph{unpatchify} to the image plane:
\begin{equation}
\label{eq:decode}
\hat{X} = \text{Unpatchify}(Z_c W_x + b_x), \qquad \hat{N} = \text{Unpatchify}(Z_n W_n' + b_n'),
\end{equation}
where \text{Unpatchify} reshapes patch predictions back to full resolution. To prevent cross-branch leakage (textures encoded as ``noise'' or noise contaminating content), we impose an orthogonality constraint between token-wise content/noise subspaces:
\begin{equation}
\label{eq:ortho}
\mathcal{L}_{\text{ortho}}
= \frac{1}{BN}\sum_{b=1}^{B}\sum_{i=1}^{N}
\left|
\left\langle
\frac{z^{(b,i)}_c}{\|z^{(b,i)}_c\|_2},
\frac{z^{(b,i)}_n}{\|z^{(b,i)}_n\|_2}
\right\rangle
\right|.
\end{equation}
Minimizing \eqref{eq:ortho} acts as a firewall against content-noise leakage. To prevent degenerate solutions (e.g., collapsing $Z_n$), we \emph{anchor} the nuisance mechanism with explicit noise supervision. For paired training with known clean target $X$, the ground-truth noise is
\begin{equation}
\label{eq:gt_noise}
N_{\text{gt}} = Y - X.
\end{equation}

\subsubsection{Teacher-guided causal prior.}
Denoising is ill-posed: multiple clean images can explain the same noisy observation. To improve
identifiability and perceptual fidelity, we leverage a \textbf{Google Nano Banana Pro (NBP)-guided causal
prior}. During training, for some samples we obtain an auxiliary teacher image $X^T$ by applying NBP in an
image-to-image editing setting with a denoising/clean-up prompt. Prior
evaluations suggest that NBP often produces high-quality, visually pleasing restorations in a zero-shot manner \cite{zuo2025nanobanana}. Since NBP outputs may contain plausible but input-inconsistent details, we adopt $X^T$ as a \emph{perceptual} target and enforce feature alignment with a fixed VGG extractor $\phi(\cdot)$:
\begin{equation}
\label{eq:teacher}
\mathcal{L}_{\text{teacher}} = \|\phi(\hat{X}) - \phi(X^T)\|_1.
\end{equation}
This distills semantic/texture priors from NBP \cite{hinton2015distill}, pulling $\hat{X}$ toward the natural-image manifold while remaining faithful to the input content. 

\subsection{Optimization and Inference}
We use the Charbonnier penalty $\rho(a)=\sqrt{a^2+\epsilon^2}$ for pixel-level supervision. The overall objective is condensed as follows:
\begin{equation}
\label{eq:loss_total}
\begin{aligned}
\mathcal{L} &= \mathcal{L}_{\text{rec}} + \lambda_{\text{noise}}\mathcal{L}_{\text{noise}} + \lambda_{\text{ortho}}\mathcal{L}_{\text{ortho}} + \lambda_{\text{teacher}}\mathcal{L}_{\text{teacher}}, \\
\mathcal{L}_{\text{rec}} &= \frac{1}{|\Omega|}\sum_{u\in\Omega}\rho\!\big(\hat{X}(u)-X(u)\big), \quad \mathcal{L}_{\text{noise}} = \frac{1}{|\Omega|}\sum_{u\in\Omega}\rho\!\big(\hat{N}(u)-N_{\text{gt}}(u)\big),
\end{aligned}
\end{equation}
where $\lambda_{\text{noise}},\lambda_{\text{ortho}},\lambda_{\text{teacher}}$ balance the interventions. When $X^T$ is unavailable, we set $\mathcal{L}_{\text{teacher}}=0$.

Thanks to patch tokenization and dynamic positional interpolation, TCD-Net supports variable resolutions. For high-resolution images where full attention becomes memory-intensive, we adopt an overlap-tile inference with Gaussian blending. We split $Y$ into overlapping patches $\{Y_k\}$ (patch size $P$, stride $S<P$), predict $\hat{X}_k=\mathcal{F}_\theta(Y_k)$, and blend them via a Gaussian weight map $W$ whose standard deviation is set to $P/4$:
\begin{equation}
\label{eq:tiled_blend}
\hat{X} = \frac{\sum_k W \odot \hat{X}_k}{\sum_k W + \varepsilon}.
\end{equation}
This reduces blocking artifacts at tile boundaries and stabilizes performance under resolution changes.

\section{Experiments}

\begin{table}[t]
\centering
\caption{Color Gaussian denoising results (PSNR, dB) on standard benchmarks at $\sigma\in\{15,25,50\}$. Best and second best are highlighted.}
\label{tab:gaussian_main}
\scriptsize
\setlength{\tabcolsep}{2.9pt}
\renewcommand{\arraystretch}{1.08}
\resizebox{0.98\textwidth}{!}{%
\begin{tabular}{l c c c c c c c c c c c c c}
\toprule
\multicolumn{2}{c}{} & \multicolumn{3}{c}{CBSD68} & \multicolumn{3}{c}{Kodak24} & \multicolumn{3}{c}{McMaster} & \multicolumn{3}{c}{Urban100} \\
\cmidrule(lr){3-5}\cmidrule(lr){6-8}\cmidrule(lr){9-11}\cmidrule(lr){12-14}
Method & Venue
& $\sigma{=}15$ & $\sigma{=}25$ & $\sigma{=}50$
& $\sigma{=}15$ & $\sigma{=}25$ & $\sigma{=}50$
& $\sigma{=}15$ & $\sigma{=}25$ & $\sigma{=}50$
& $\sigma{=}15$ & $\sigma{=}25$ & $\sigma{=}50$ \\
\midrule
DnCNN & TIP'17 & 33.90 & 31.24 & 27.95 & 34.60 & 32.14 & 28.95 & 33.45 & 31.52 & 28.62 & 32.98 & 30.81 & 27.59 \\
DRUNet & TPAMI'21 & 34.30 & 31.69 & 28.51 & 35.31 & 32.89 & 29.86 & 35.40 & 33.14 & 30.08 & 34.81 & 32.60 & 29.61 \\
SwinIR & ICCV'21 & 34.42 & 31.78 & 28.56 & 35.34 & 32.89 & 29.79 & 35.61 & 33.20 & 30.22 & 35.13 & 32.90 & 29.82 \\
SCUNet & AAAI'22 & 34.40 & \underline{31.79} & 28.61 & 35.34 & 32.92 & 29.87 & 35.60 & \underline{33.34} & \underline{30.29} & 35.18 & 33.03 & 30.14 \\
Restormer & CVPR'22 & 34.39 & 31.78 & 28.59 & \underline{35.44} & 33.02 & 30.00 & 35.55 & 33.31 & \underline{30.29} & 35.06 & 32.91 & 30.02 \\
NAFNet & ECCV'22 & $-$ & $-$ & 28.39 & $-$ & $-$ & 29.64 & $-$ & $-$ & 29.98 & $-$ & $-$ & 29.71 \\
HAT & CVPR'23 & 34.42 & \underline{31.79} & 28.58 & \textbf{35.46} & \underline{33.03} & \underline{30.01} & \underline{35.62} & \underline{33.34} & 30.26 & \textbf{35.37} & \underline{33.14} & \underline{30.23} \\
MambaIR & ECCV'24 & 34.31 & 31.66 & 28.47 & 35.33 & 32.82 & 29.94 & 35.40 & 33.15 & 30.06 & 35.17 & 32.99 & 30.07 \\
MambaIRv2 & CVPR'25 & \textbf{34.44} & 31.78 & \textbf{28.66} & \underline{35.44} & \textbf{33.05} & \textbf{30.02} & 35.55 & 33.31 & 30.28 & 35.27 & \underline{33.14} & 30.21 \\
\textbf{TCD-Net} & $-$ & \underline{34.43} & \textbf{31.80} & \underline{28.64} & 35.43 & 32.99 & 30.00 & \textbf{35.64} & \textbf{33.35} & \textbf{30.34} & \underline{35.35} & \textbf{33.16} & \textbf{30.27} \\
\bottomrule
\end{tabular}}%
\end{table}

\begin{figure}[t]
  \centering
  \includegraphics[width=0.95\textwidth]{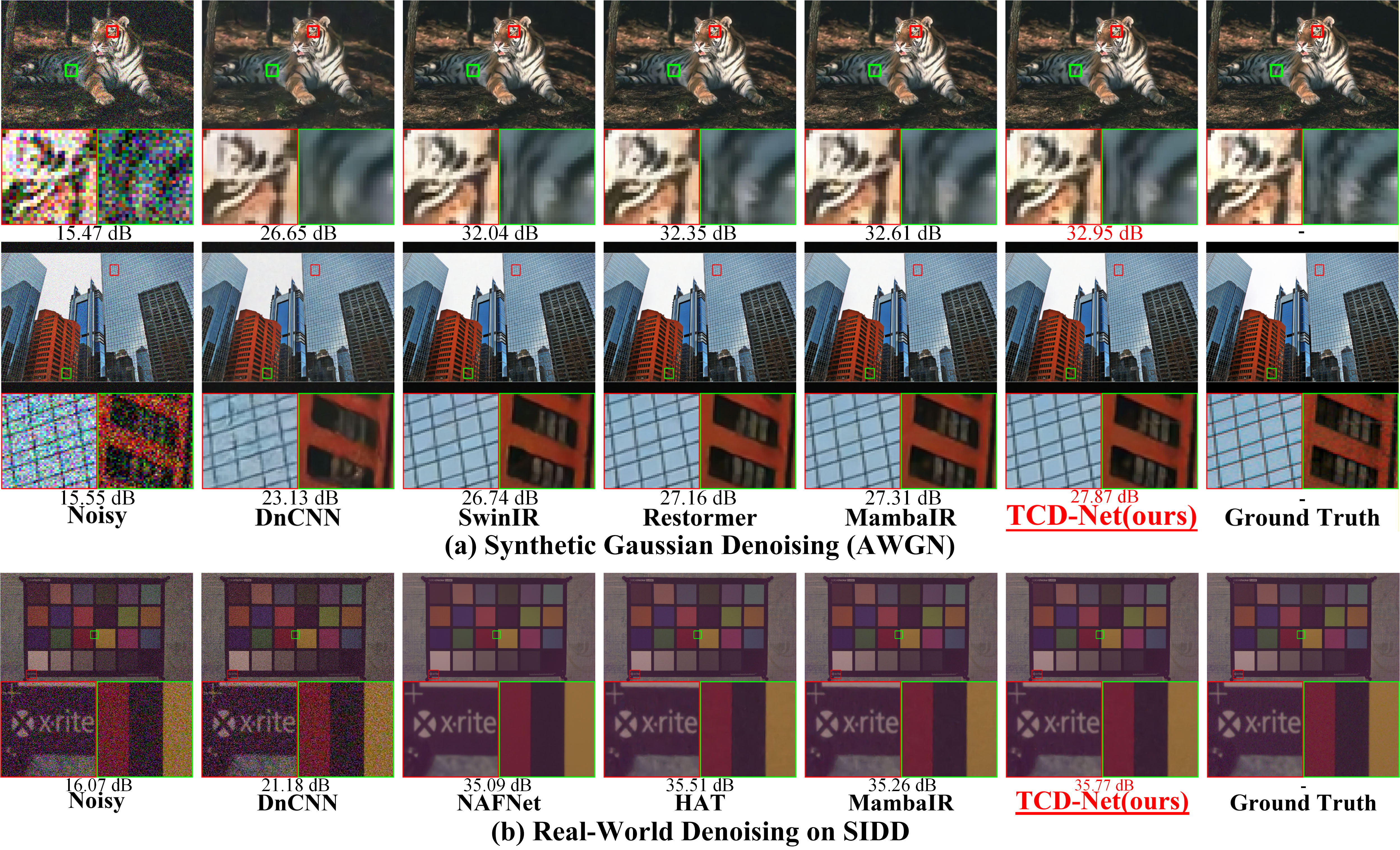}
  \caption{\textbf{Qualitative comparison on synthetic and real noise.}
  \textbf{(a) Synthetic Gaussian denoising (AWGN).}
  \textbf{(b) Real-world denoising on SIDD.}
  Conventional methods may leave residual noise or oversmooth details due to spurious content--noise correlation.
  With EBA-based intervention, orthogonal disentanglement, and teacher guidance, TCD-Net restores cleaner results with sharper textures and edges.}
  \label{fig:demo}
\end{figure}

\begin{table}[t]
\centering
\caption{\textbf{Real-world denoising} on SIDD and DND (PSNR/SSIM). We fine-tune the DF2K pretrained model on SIDD and DND, respectively. Best and second best are highlighted.}
\label{tab:real_results}
\scriptsize
\setlength{\tabcolsep}{2.4pt}
\renewcommand{\arraystretch}{1.05}
\resizebox{0.98\textwidth}{!}{%
\begin{tabular}{l c c c c c c c c c c c c c c c}
\toprule
Dataset & Metric &
\shortstack{BM3D\\TIP'07} &
\shortstack{DnCNN\\TIP'17} &
\shortstack{CBDNet\\CVPR'19} &
\shortstack{RIDNet\\ICCV'19} &
\shortstack{MPRNet\\CVPR'21} &
\shortstack{DAGL\\ICCV'21} &
\shortstack{SwinIR\\ICCV'21} &
\shortstack{Uformer\\CVPR'22} &
\shortstack{Restormer\\CVPR'22} &
\shortstack{NAFNet\\ECCV'22} &
\shortstack{HAT\\CVPR'23} &
\shortstack{MambaIR\\ECCV'24} &
\shortstack{MambaIRv2\\CVPR'25} &
\shortstack{\textbf{TCD-Net}\\(ours)} \\
\midrule
SIDD & PSNR & 25.65 & 23.66 & 30.78 & 38.71 & 39.71 & 38.94 & 40.05 & 39.77 & 40.02 & 40.30 & \underline{40.46} & 39.89 & 40.21 & \textbf{40.48} \\
    & SSIM & 0.685 & 0.583 & 0.801 & 0.951 & 0.958 & 0.953 & 0.961 & 0.959 & 0.960 & 0.962 & \underline{0.964} & 0.960 & 0.963 & \textbf{0.965} \\
DND  & PSNR & 34.51 & 32.43 & 38.06 & 39.26 & 39.80 & 39.77 & 39.91 & 39.96 & 40.03 & 40.27 & \underline{40.44} & $-$ & $-$ & \textbf{40.45} \\
    & SSIM & 0.851 & 0.790 & 0.942 & 0.953 & 0.954 & 0.956 & \textbf{0.961} & 0.956 & 0.956 & 0.957 & 0.958 & $-$ & $-$ & \underline{0.960} \\
\bottomrule
\end{tabular}}%
\end{table}

\begin{table}[htbp]
  \centering
  \begin{minipage}[t]{0.42\textwidth}
    \centering
    \caption{Perceptual quality comparison using LPIPS (lower is better) on SIDD and Urban100 ($\sigma{=}50$).}
    \label{tab:lpips}
    \vspace{0.1em}
    \resizebox{\linewidth}{!}{% 
      \begin{tabular}{l c c c}
      \toprule
      Method & Venue & SIDD & Urban100\\
            &       &      & ($\sigma{=}50$) \\
      \midrule
      SwinIR & ICCV'21 & 0.142 & 0.210 \\
      Uformer & CVPR'22 & 0.140 & 0.205 \\
      Restormer & CVPR'22 & 0.138 & 0.198 \\
      NAFNet & ECCV'22 & 0.135 & 0.195 \\
      HAT & CVPR'23 & 0.131 & 0.188 \\
      MambaIR & ECCV'24 & 0.125 & 0.182 \\
      MambaIRv2 & CVPR'25 & 0.118 & 0.174 \\
      \textbf{TCD-Net} & $-$ & 0.128 & 0.184 \\
      \bottomrule
      \end{tabular}%
    }
  \end{minipage}
  \hfill
  \begin{minipage}[t]{0.54\textwidth}
    \centering
    \caption{Efficiency comparison. FLOPs are reported for a single forward pass. Latency/FPS are measured at $256\times256$ with batch size 1. Best and second best are highlighted.}
    \label{tab:efficiency}
    \vspace{0.1em}
    \resizebox{\linewidth}{!}{% 
      \begin{tabular}{@{}l c r r r c c@{}}
      \toprule
      Method & Venue &
      \shortstack{FLOPs\\(G)} &
      \shortstack{Time\\(ms)} &
      FPS &
      \shortstack{Gauss.\\PSNR} &
      \shortstack{Real\\PSNR} \\
      \midrule
      SwinIR & ICCV'21 & 804.7 & 154.5 & 6.5 & 34.42 & 40.05 \\
      Uformer & CVPR'22 & 86.9 & 18.6 & 53.7 & 34.32 & 39.77 \\
      Restormer & CVPR'22 & 154.9 & 30.2 & 33.1 & 34.39 & 40.02 \\
      NAFNet & ECCV'22 & \textbf{16.3} & \underline{10.2} & \underline{98.5} & 34.35 & 40.30 \\
      HAT & CVPR'23 & 721.3 & 258.8 & 3.9 & 34.42 & \underline{40.46} \\
      MambaIR & ECCV'24 & \underline{68.2} & 22.1 & 45.2 & 34.31 & 39.89 \\
      MambaIRv2 & CVPR'25 & 69.5 & 18.1 & 55.4 & \textbf{34.44} & 40.21 \\
      \textbf{TCD-Net} & $-$ & 84.8 & \textbf{9.6} & \textbf{104.2} & \underline{34.43} & \textbf{40.48} \\
      \bottomrule
      \end{tabular}%
    }
  \end{minipage}
\end{table}

\begin{table}[t]
\centering
\caption{Ablation on CBSD68 validation at $\sigma{=}25$ (PSNR, dB). We incrementally add the dual-stream design with noise supervision, the independent orthogonality regularizer, CPE, the EBA module, and the teacher prior.}
\label{tab:ablation}
\scriptsize
\setlength{\tabcolsep}{2.4pt}
\renewcommand{\arraystretch}{1.05}
\resizebox{0.85\columnwidth}{!}{%
\begin{tabular}{@{}c c c c c c c@{}}
\toprule
ID &
\shortstack{Dual-Stream\\$\&~\mathcal{L}_{\text{noise}}$} &
\shortstack{$\mathcal{L}_{\text{ortho}}$\\(Indep.)} &
\shortstack{Pos.\\Enc.} &
\shortstack{EBA \\Topology} &
\shortstack{$\mathcal{L}_{\text{teacher}}$\\(Prior)} &
PSNR \\
\midrule
1 & $\times$ & $\times$ & APE & $\times$ & $\times$ & 31.35 \\
2 & \checkmark & $\times$ & APE & $\times$ & $\times$ & 31.36 \\
3 & \checkmark & \checkmark & CPE & $\times$ & $\times$ & 31.48 \\
4 & \checkmark & \checkmark & CPE & Parallel & $\times$ & 31.60 \\
5 & \checkmark & \checkmark & CPE & Serial & $\times$ & 31.68 \\ 
6 & \checkmark & \checkmark & CPE & Serial & \checkmark & 31.80 \\
\bottomrule
\end{tabular}}%
\end{table}

\subsection{Experimental Setup}
\subsubsection{Tasks and benchmarks.}
We evaluate \textbf{TCD-Net} on (i) \emph{synthetic} color Gaussian denoising and (ii) \emph{real-world} denoising.
For synthetic denoising, we follow the standard blind AWGN protocol \cite{zhang2017dncnn,zhang2018ffdnet}: during training, $\sigma$ is uniformly sampled from $[0,50]$ and added to clean RGB patches; during testing, we report PSNR (dB) at $\sigma\in\{15,25,50\}$ on CBSD68 \cite{zhang2017dncnn}, Kodak24, McMaster, and Urban100 \cite{Huang_2015_CVPR}. 
For real-world denoising, we report PSNR/SSIM on SIDD \cite{abdelhamed2018sidd} and DND \cite{plotz2017dnd}, and additionally report LPIPS \cite{zhang2018lpips} (lower is better) on SIDD and on Urban100 at $\sigma{=}50$. 
All real-noise results are obtained by \emph{fine-tuning} a synthetic-pretrained model: we start from a DF2K-trained Gaussian denoiser (DIV2K+Flickr2K) \cite{agustsson2017ntire,Lim_2017_CVPR_Workshops} and fine-tune on SIDD and DND following their training protocols.

\subsubsection{Implementation and training.}
Unless specified otherwise, we use a ViT-L/16 backbone \cite{dosovitskiy2021vit} (dim $1024$, depth $24$, $16$ heads) with the causal interventions in Section~\ref{sec:method}.
We train on DF2K \cite{agustsson2017ntire,Lim_2017_CVPR_Workshops} using $256\times256$ crops and standard augmentations. We optimize with AdamW (weight decay $1\mathrm{e}{-4}$) and cosine learning-rate decay.
We adopt a two-stage schedule: \emph{Stage~1} trains for 100 epochs (lr $2\mathrm{e}{-4}$, batch 32, repetition 20) with reconstruction, noise supervision, and orthogonality ($\lambda_{\text{noise}}{=}0.5$, $\lambda_{\text{ortho}}{=}0.1$); \emph{Stage~2} fine-tunes for 50 epochs (lr $5\mathrm{e}{-5}$) enabling teacher guidance (weight $0.1$) and slightly reducing regularization ($\lambda_{\text{noise}}{=}0.25$, $\lambda_{\text{ortho}}{=}0.05$).
For \textbf{real-world} denoising, we initialize from the DF2K-trained model and \textbf{fine-tune on SIDD/DND} with the same objectives (Eq.~\eqref{eq:loss_total}), disabling synthetic noise injection and using dataset-provided noisy/clean pairs for supervision \cite{abdelhamed2018sidd,plotz2017dnd}.

\subsubsection{Hardware and software.}
All experiments are implemented in PyTorch~2.8.0 with Python~3.12 on Ubuntu~22.04, using CUDA~12.8. Runtime measurements are conducted on a single RTX~5090 (32GB). Latency is measured with batch size 1 at resolution $256\times256$ after warm-up, and FPS is computed as the reciprocal of mean latency.

\subsubsection{Baselines.}
We compare with a classical prior (BM3D~\cite{dabov2007bm3d}), representative CNN denoisers (DnCNN~\cite{zhang2017dncnn}, CBDNet~\cite{guo2019cbdnet}, RIDNet~\cite{anwar2019ridnet}, DRUNet~\cite{zhang2021drunet}), and modern Transformer/state-space restorers (SwinIR~\cite{liang2021swinir}, DAGL~\cite{mou2021dagl}, MPRNet~\cite{zamir2021mprnet}, SCUNet~\cite{zhang2023scunet}, Uformer~\cite{wang2022uformer}, Restormer~\cite{zamir2022restormer}, NAFNet~\cite{chen2022nafnet}, HAT~\cite{chen2023hat}, MambaIR~\cite{guo2024mambair}, MambaIRv2~\cite{guo2025mambairv2}). We follow standard benchmark protocols and report results from official implementations or widely adopted reported numbers.

\subsection{Results on Synthetic Gaussian Denoising}
Table~\ref{tab:gaussian_main} reports PSNR performance of TCD-Net against recent restorers on four benchmarks. Overall, TCD-Net achieves \emph{consistently strong} performance and excels on texture-rich datasets, obtaining the \textbf{best} PSNR on McMaster for all $\sigma\in\{15,25,50\}$ and on Urban100 at $\sigma{=}25$/$50$ (runner-up at $\sigma{=}15$), while remaining top-tier on CBSD68 across noise levels. The gains under heavy noise (e.g., $\sigma{=}50$) align with our causal design: explicit noise supervision provides an identifiable nuisance handle, and the orthogonality constraint reduces content--noise cross-talk that causes over-smoothing or residual noise-like details. Moreover, CPE alleviates resolution-induced representation shifts and EBA mitigates environment-related bias, improving stability across settings. Visual comparisons in Fig.~\ref{fig:demo} further demonstrate that TCD-Net restores cleaner results with sharper textures. Together with Table~\ref{tab:efficiency}, TCD-Net provides a practical quality--speed trade-off by emphasizing robustness via explicit mechanism separation rather than backbone scaling.

\subsection{Real-World Adaptation, Efficiency, and Ablations}

\subsubsection{Real-world results: fine-tuning from the Gaussian model.}
Table~\ref{tab:real_results} reports performance on real noisy photographs.
Starting from the DF2K-trained Gaussian model and fine-tuning on SIDD/DND, TCD-Net achieves the \textbf{best} PSNR/SSIM on SIDD and the \textbf{best} PSNR on DND, indicating strong synthetic-to-real transfer.
Notably, it also attains competitive DND SSIM (runner-up), suggesting that the explicit nuisance branch and the de-confounding components help adapt to signal-dependent noise and camera pipeline variations during fine-tuning.

\subsubsection{Perceptual quality: LPIPS.}
Besides distortion metrics, we report LPIPS~\cite{zhang2018lpips} on SIDD and Urban100 ($\sigma{=}50$) to better reflect perceptual similarity.
Table~\ref{tab:lpips} reports LPIPS (lower is better) for perceptual comparison.
TCD-Net achieves competitive perceptual quality (e.g., 0.128 on SIDD and 0.184 on Urban100 at $\sigma{=}50$), outperforming several strong Transformer baselines, while the best LPIPS is obtained by the latest SSM-based restorers.
This suggests that our explicit content--noise separation preserves perceptual fidelity, though there remains room to further improve perceptual alignment.

\subsubsection{Efficiency: fast inference with a simple, single-path Transformer.}
Beyond restoration fidelity, we emphasize that TCD-Net is designed to be efficient at inference.
As shown in Table~\ref{tab:efficiency}, TCD-Net achieves the \textbf{lowest latency} (9.59\,ms) and the \textbf{highest FPS} (104.2) among all compared methods on our hardware, while maintaining strong denoising quality.
Although its FLOPs are not the smallest, the computation graph is intentionally simple and straight-through, which favors GPU parallelism and reduces synchronization overhead in practice.

\subsubsection{Ablation: effects of causal components and teacher prior.}
Table~\ref{tab:ablation} shows that the proposed modules contribute \emph{complementarily}. A naive dual-head yields only a marginal gain (ID~1$\rightarrow$2, +0.01\,dB), indicating that architectural splitting alone does not guarantee disentanglement. Adding the orthogonality regularizer and switching to CPE brings a larger improvement (ID~2$\rightarrow$3, +0.12\,dB), suggesting reduced content--noise leakage and more stable dense prediction under positional shifts. Introducing EBA further boosts PSNR (ID~3$\rightarrow$4/5), and the serial topology outperforms the parallel one (ID~4$\rightarrow$5, \textbf{+0.08\,dB}). Finally, enabling the NBP teacher prior provides an additional gain (ID~5$\rightarrow$6, \textbf{+0.12\,dB}), consistent with improved identifiability by pulling the content prediction toward the natural-image manifold; this teacher loss is applied only during training and incurs no inference cost. 

\section{Conclusion}
We revisit image denoising from a causal-intervention perspective, addressing how correlation-based fitting entangles content and noise. We propose TCD-Net, a Vision Transformer denoiser with structured interventions for content--noise disentanglement: EBA deconfounds environment bias via de-centered projection with residual restoration, and a dual-branch head with orthogonality plus strong noise supervision reduces leakage and anchors the nuisance factor. We further distill an NBP-guided feature-level perceptual prior to regularize content toward the natural-image manifold, and adopt resolution-stable positional encoding for better generalization. Experiments on synthetic Gaussian and real benchmarks show state-of-the-art or competitive fidelity with high efficiency (9.59\,ms, 104.2 FPS on RTX~5090). Future work will explore reliability-aware teacher priors and causal intervention learning under domain shift and weak supervision for real-world noise.

\bibliographystyle{splncs04}
\bibliography{references}

\end{document}